\documentclass[runningheads]{llncs}

\usepackage{eccv}
\usepackage{eccvabbrv}
\usepackage{array}
\usepackage{graphicx}
\usepackage{booktabs}
\usepackage{caption}
\usepackage[accsupp]{axessibility}
\usepackage{hyperref}
\usepackage[table]{xcolor}
\usepackage{orcidlink}

\begin{document}

\title{DocCogito: Aligning Layout Cognition and Step-Level Grounded Reasoning for Document Understanding}

\titlerunning{DocCogito for Document Understanding}

\author{
Yuchuan Wu \and
Minghan Zhuo \and
Teng Fu \and
Mengyang Zhao \and
Bin Li \textsuperscript{*} \and
Xiangyang Xue
}

\authorrunning{Y. Wu et al.}

\institute{
Fudan University, Shanghai, China \\
\email{\{ycwu24, mhzhuo24, tfu23\}@m.fudan.edu.cn} \\
\email{\{myzhao20, libin, xyxue\}@fudan.edu.cn}
}

\maketitle

\begingroup
\renewcommand\thefootnote{}
\footnotetext{* Corresponding author.}
\endgroup

\begin{abstract}
Document understanding with multimodal large language models (MLLMs) requires not only accurate answers but also explicit, evidence-grounded reasoning, especially in high-stakes scenarios.
However, current document MLLMs still fall short of forming a complete, human-like reasoning process, because even when they improve both layout encoding and CoT-style prompting, the interaction between the two is typically learned implicitly and remains loosely coupled rather than being enforced as a systematic mechanism.
So we propose \textbf{DocCogito}, a unified framework that integrates global layout perception with structured, region-grounded reasoning.
DocCogito introduces a lightweight layout tower that distills page structure into learnable \emph{global layout prior tokens}, and a deterministic \emph{Visual--Semantic Chain} (VSC)—a concise structured representation less ambiguous than free-form natural-language CoT—to supervise fine-grained intermediate reasoning aligned with evidence regions.
Training follows a progressive recipe, including layout perception pretraining, VSC-guided cold start, rejection sampling, and GRPO.
To further strengthen the internal coupling between layout priors and VSC execution, we augment standard rewards with a fine-grained region-confidence signal that encourages reasoning traces to stay aligned with corresponding evidence regions.
Extensive experiments on six benchmarks (DocVQA, WTQ, ChartQA, TextVQA, OCRBench, and InfoVQA) demonstrate strong generalization, achieving state-of-the-art results on four benchmarks.

\end{abstract}

\section{Introduction}
\label{fig:cot_paradigm}

\begin{figure}
    \centering
   \includegraphics[width=0.95\linewidth]{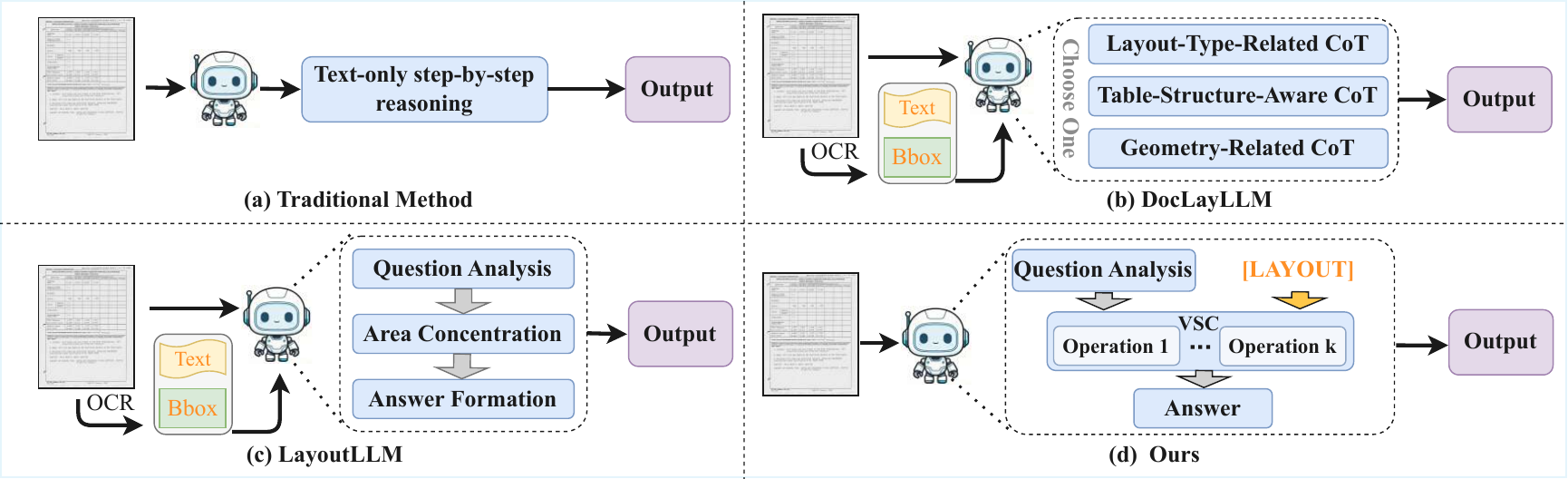}
    \caption{Overview of reasoning pipelines in document understanding models. (a) Traditional text-only CoT. (b) DocLayLLM with OCR-based text/boxes and multiple CoT templates. (c) LayoutLLM with OCR-based text/boxes and staged reasoning (question analysis $\rightarrow$ area concentration $\rightarrow$ answer formation). (d) Ours, an OCR-free approach that integrates global layout perception with structured, region-grounded reasoning. VSC means Visual-Semantic Chain.}
   \label{figure1}
\end{figure}

Document understanding~\cite{feng2024docpedia,wang2025marten,appalaraju2024docformerv2,huang2024mini,liu2024hrvda,park2024hierarchical} aims to extract, reason over, and answer questions about heterogeneous elements in visually rich documents, including text blocks, tables, charts, and figures.
With the rapid digitization of paper-based materials and the increasing adoption of intelligent office systems, document understanding powered by multimodal large language models (MLLMs)~\cite{yu2024texthawk2,hu2025mplug,chen2024internvl,li2024monkey,zhang2025dockylin,liao2025doclayllm,yu2025docthinker} has become a central research direction.
Moreover, in high-stakes applications such as legal contracts and financial reports, these demands impose stricter requirements on document understanding models: beyond answer accuracy, they must provide explicit, grounded reasoning with precise evidence localization.

Despite its importance, document MLLMs rarely demonstrate the complete reasoning behavior like humans do. Human readers first establish a global layout prior and then iteratively gather evidence and compose simple operations to reach the answer. This blueprint is beneficial not for human-likeness per se, but because it provides a transferable inductive bias for generalization—global browsing narrows the search space under layout shifts, while step-wise evidence use supports compositional reasoning and mitigates distraction-induced errors.

Following this human blueprint, recent document-oriented MLLMs have indeed advanced (i) layout modeling for evidence navigation and (ii) controllable reasoning via chain-of-thought (CoT) (Fig.~\ref{figure1}).
DocLayLLM~\cite{liao2025doclayllm} further incorporates disentangled layout cues together with multiple CoT templates; LayoutLLM~\cite{luo2024layoutllm} combines OCR-based layout-aware encoding with a staged LayoutCoT.
However, these efforts mainly strengthen spatial encoding or prompt-level controllability in isolation, rather than enforcing a coupled process.
Specifically, layout information is not distilled into a direct prior within the CoT process to reliably guide question-region localization under layout shifts; nor is the reasoning steps constrained so that each intermediate step continuously focuses on the corresponding evidence region while composing operations.
As a result, models can still drift to distractors or shortcut with free-form rationales when layouts vary.

To tackle these issues and enable a complete, human-like reasoning process, we propose \textbf{DocCogito}, an OCR-free unified framework that integrates global layout perception and structured, region-grounded reasoning via progressive training recipe.
DocCogito introduces a lightweight layout tower to encode page structure into an explicit global layout prior (as learnable layout tokens), and a deterministic VSC, a concise structured representation that is shorter and less semantically ambiguous than natural-language CoT, to supervise fine-grained, evidence-region-aligned intermediate reasoning.
Training follows a progressive curriculum: we first pretrain layout perception, and then conduct multi-phase post-training with VSC-guided cold start, supervised fine-tuning, and GRPO to refine reasoning behaviors.
To further strengthen the internal coupling between the layout prior and VSC execution, we augment standard reward signals with a fine-grained region-confidence term, encouraging reasoning traces to remain aligned with their corresponding evidence regions.

We evaluate DocCogito on six benchmarks, including DocVQA~\cite{mathew2021docvqa}, WTQ~\cite{pasupat2015compositional}, ChartQA~\cite{masry2022chartqa}, TextVQA~\cite{Singh_2019_CVPR}, OCRBench~\cite{liu2024ocrbench}, and InfoVQA~\cite{mathew2022infographicvqa}.
DocCogito achieves strong performance across benchmarks and reaches state of the art on four benchmarks.

Our main contributions are as follows:
\begin{itemize}
    \item We propose \textbf{DocCogito}, an OCR-free unified document MLLM framework that couples global layout perception with step-wise, region-grounded execution to enable a complete, human-like reasoning process.
    \item We introduce a lightweight layout tower that distills page structure into explicit \emph{global layout prior tokens}, and a deterministic \emph{VSC} to supervise fine-grained, evidence-region-aligned intermediate reasoning.
    \item We develop a progressive training recipe (layout perception pretraining $\rightarrow$ VSC-guided cold start $\rightarrow$ rejection sampling $\rightarrow$ GRPO) and augment standard rewards with a fine-grained region-confidence signal to strengthen the coupling between layout priors and VSC execution.
    \item DocCogito achieves comparable and even state-of-the-art performance across six benchmarks, demonstrating strong generality and scalability across model sizes, and ablation studies further validate the effectiveness of DocCogito.
\end{itemize}
\section{Related Work}
\label{sec:related_work}

\subsection{Document Understanding}

Multimodal large language models (MLLMs)~\cite{appalaraju2024docformerv2,huang2024mini,liu2024hrvda,park2024hierarchical,yu2024texthawk2,hu2025mplug,chen2024internvl,li2024monkey,zhang2025dockylin,gu2021unidoc,xu2021layoutlmv2} have demonstrated strong capability in document understanding by jointly modeling textual content, layout structure, and visual appearance. Recent advances enhance layout awareness and spatial reasoning through layout-aware encoders and chain-of-thought (CoT)~\cite{wei2022chain} prompting. For example, DocLayLLM~\cite{liao2025doclayllm} integrates visual patch tokens with 2D positional embeddings to strengthen structural grounding, while DocPedia~\cite{feng2024docpedia} explores alternative representations to better preserve visual--text correlations. Models such as Marten~\cite{wang2025marten} further improve visual grounding through region-aware attention mechanisms.

Beyond layout encoding, several approaches incorporate reinforcement learning to refine reasoning behaviors. DocThinker~\cite{yu2025docthinker} employs rule-based feedback for explainable CoT planning, and Point-RFT~\cite{ni2025point} leverages visually grounded reinforcement tuning to improve multimodal coherence. Collectively, these works illustrate a growing trend toward modular, interpretable, and layout-aware MLLMs driven by CoT supervision or adaptive optimization. However, existing methods still focus primarily on surface-level structural cues and lack comprehensive cognitive modeling that jointly aligns layout perception with reasoning processes.



\subsection{Reinforcement Learning}

Reinforcement learning (RL) has made significant progress in improving structured reasoning, planning, and alignment in multimodal large language models (MLLMs). Early studies such as RLAIF~\cite{lee2023rlaif} demonstrate that reward-driven optimization can complement or even replace supervised fine-tuning by providing dynamic, preference-based signals. The Explore–Execute Chain framework~\cite{yang2025explore} further shows that separating exploration and execution enables more stable multi-step reasoning and structure-aware planning. In particular, GRPO~\cite{shao2024deepseekmath} introduces a gradient-based preference optimization scheme that enables stable policy updates without requiring explicit reward modeling, making it especially effective for long-horizon and multi-step reasoning.
Recent work extends RL to long-horizon reasoning through verifier-guided reward shaping~\cite{huang2025loong}, self-refinement mechanisms~\cite{madaan2023self}, and preference optimization methods such as ORPO~\cite{hong2024orpo} and DPO~\cite{hong2024orpo,rafailov2023direct}, which improve consistency and controllability of generated reasoning traces. These methods move beyond static Chain-of-Thought (CoT) prompting~\cite{wei2022chain} by enabling adaptive reasoning trajectories, dynamic error correction, and reward-aligned behavioral control. 

In parallel, agent-style approaches such as ReAct-based RL fine-tuning~\cite{yao2022react} highlight that RL can support interpretable, task-aware decision making in more complex reasoning settings.
Collectively, these developments position RL as a flexible and powerful paradigm for aligning model behavior with downstream reasoning objectives. However, existing RL methods are primarily designed for textual reasoning and lack integration with layout-grounded cognitive signals, motivating our GRPO-based approach with visually grounded rewards.

\section{Methodology}

\begin{figure*}
    \centering
   \includegraphics[width=1\linewidth]{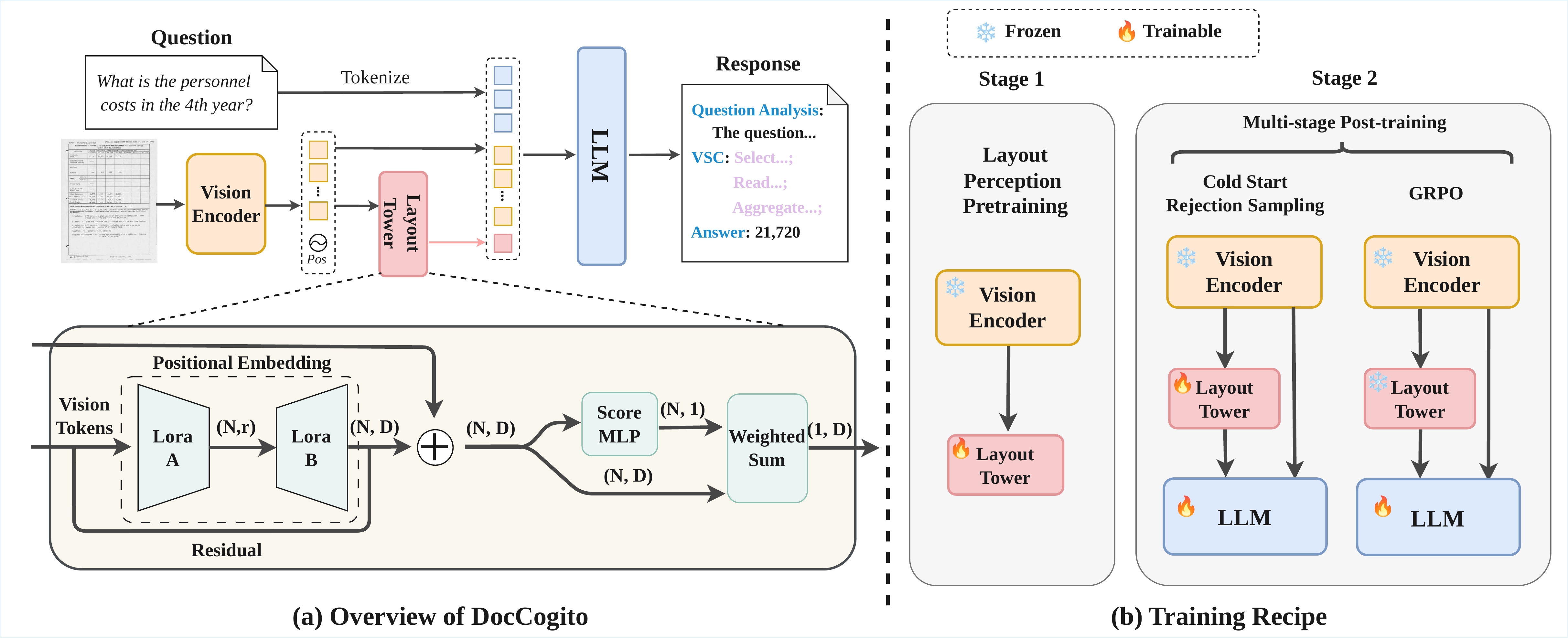}

   \caption{Overview of our methods. (a) The model architecture, where a lightweight layout tower extracts global layout cues and injects a \texttt{[LAYOUT]} token into the LLM. 
(b) The two-stage training recipe, consisting of layout perception pretraining and multi-stage GRPO.}
   \label{figure2}
\end{figure*}

\subsection{Dataset Construction}
\label{sec:Dataset Construction}

We construct a stage-aligned training corpus tailored to DocCogito. It consists of a layout-focused pretraining corpus for layout tower pretraining, a VSC-style CoT dataset for structured reasoning alignment, and a multi-domain QA corpus that provides diverse trajectories for rejection sampling and GRPO. 


\textbf{Layout-focused Pretraining Corpus.}
To initialize the layout tower with explicit page-structure awareness, we construct a layout-focused pretraining corpus using 20,000 OCR-annotated samples from DocVQA~\cite{mathew2021docvqa}, covering diverse formats, \eg, forms, invoices, and reports, \etal. Each sample contains a page containing text-line and region-level bounding boxes that provide fine-grained spatial annotations reflecting the document layout. These OCR coordinates supply the structural cues required for learning page-level layout priors.


\textbf{VSC-style CoT Dataset.}
To support structured, layout-grounded reasoning, we construct a VSC-style CoT dataset in which each sample follows the three-part format of $\langle \texttt{question\_analysis}, \texttt{vsc},  \texttt{answer}\rangle$. The full design of vsc is described in Section~\ref{vsc}.
We curate a cold-start corpus of 4,000 samples from nine categories of DocVQA, covering common document layouts such as forms, tables, lists, and paragraphs. Each instance is annotated using Qwen3-VL~\cite{yang2025qwen3} to generate the VSC-formatted reasoning sequences and corresponding answers. All data will subsequently undergo expert review. This structured dataset provides clean, layout-aware supervision that prepares the model for subsequent large-scale reasoning refinement.



\textbf{Multi-Domain QA Corpus.}
To enhance the model's robustness and generalization to different data types, we construct a unified multi-domain QA corpus by integrating samples from DocVQA~\cite{mathew2021docvqa}, InfographicsVQA~\cite{mathew2022infographicvqa}, OCRVQA~\cite{mishra2019ocr}, and TextVQA~\cite{Singh_2019_CVPR}. Following a balanced domain allocation strategy, we subsample roughly 100k samples, removing duplicated images, and filtering low-quality OCR outputs. Finally, the collected corpus ensures that each sample can be expressed in the VSC triplet format.
To avoid skewing toward OCR-heavy domains, we adopt a sampling ratio of \textbf{$6:4:5:5$}, balancing different types of samples.




\subsection{Visual-Semantic Chain (VSC)}
\label{vsc}

\begin{table}[t]
\centering
\caption{Operations in the Visual-Semantic Chain (VSC).}
\label{tab:vsc_ops}
\renewcommand{\arraystretch}{1.15}
\begin{tabular}{p{0.20\linewidth} >{\raggedright\arraybackslash\footnotesize}p{0.70\linewidth}}
\toprule
\textbf{Operation} & \textbf{Description} \\
\midrule
\texttt{Select} & Choose a semantic region (\eg, header, table, cell) using structural or key cues. \\
\hline
\texttt{Read} & Extract textual content from the selected region. \\
\hline
\texttt{Filter} & Filter candidates using predicates (\eg, key contains ``Revenue'', year==``2024''). \\
\hline
\texttt{Compare} & Apply comparison metrics (\textit{eq, max, min}) to identify the intended value. \\
\hline
\texttt{Aggregate} & Combine multiple values (\eg, summation or string concatenation). \\
\bottomrule
\end{tabular}
\end{table}

As illustrated in Fig.~\ref{figure1}, existing chain-of-thought (CoT) approaches for document understanding, ranging from natural-language CoT to layout-augmented templates, remain fundamentally expressed in text. Although interpretable, natural-language CoTs suffer from several inherent issues: 
differences of granularity between thinking steps, frequently implicit assumptions, 
and ambiguous linguistic descriptions.
As a result, natural-language CoT provides an unstable supervision signal for layout-sensitive reasoning.

To address these limitations, we propose the Visual–Semantic Chain (VSC), a structured representation that decomposes reasoning into atomic, layout-grounded operations. Each step is represented as a triplet:
\begin{equation}
\texttt{step} = \langle\texttt{op}, \texttt{region}, \texttt{args}\rangle,
\label{eq:vsc_triplet}
\end{equation}
where \texttt{op} denotes a primitive reasoning operator, \texttt{region} anchors the operation to a specific layout area, and \texttt{args} provides minimal, auditable parameters.

\textbf{Triplet Formulation.}
As shown in Table~\ref{tab:vsc_ops}, VSC builds upon a compact set of layout-grounded primitive operations, which form a minimal yet compositional action space for fine-grained visual--semantic reasoning.
Each thinking step is explicitly anchored to a layout region, while its arguments supply structured cues such as keys or predicates, enabling deterministic and machine-perceptible supervision.
Although compact, these five primitives are sufficient to cover all tasks in our current evaluation suite.
Fig.~\ref{fig:vis} and Fig.~\ref{fig:vis_all} provides diverse VSC-style CoT examples as an initial empirical validation, and a more detailed analysis and discussion are provided in Suppl.\ Sec.~A.






\subsection{Model Architecture}

Our framework follows the standard structure of multimodal large language models (MLLMs)~\cite{yang2025qwen3}, mainly consisting of a vision encoder for visual perception and a large language model~\cite{yang2025qwen3} for reasoning. To equip the model with the additional layout modeling capability required for reliable document reasoning, we introduce a lightweight layout tower that extracts global structural cues from document images and injects layout-aware priors into the reasoning process.
As illustrated in Fig.~\ref{figure2}, the layout tower is attached to the visual encoder and produces a compact \texttt{[LAYOUT]} token. This token is projected into the embedding and concatenated with the text embeddings, ensuring that layout priors influence all downstream cross-modal reasoning.



\textbf{Layout Tower.}
Given patch-level visual embeddings 
$\mathbf{V}=\{\mathbf{v}_1,\dots,\mathbf{v}_N\}$ 
from the vision encoder, 
the layout tower first applies LoRA-based~\cite{hu2022lora} adapter to inject layout-sensitive transformations:
\begin{equation}
\mathbf{h}_i = \mathbf{v}_i + \Delta W_{\text{LoRA}}\mathbf{v}_i,
\label{eq:lora_adapter}
\end{equation}
where $\Delta W_{\text{LoRA}}$ denotes the weights of adapter. 
The low-rank structure enables the module to model document-specific structural subspaces with minimal additional parameters.
To preserve spatial priors, positional embeddings are added to $\mathbf{h}_i$, followed by a learnable scoring module:
\begin{equation}
\alpha_i = \text{Softmax}(\text{MLP}_{\text{score}}(\mathbf{h}_i + \mathbf{h}_{pos})),
\label{eq:score_mlp}
\end{equation}
where $\alpha_i$ represents the normalized importance of each patch token, reflecting its contribution to global layout organization. Then, a global layout token is obtained via weighted addition:
\begin{equation}
\mathbf{L} = \sum_{i=1}^N \alpha_i \cdot \mathbf{h}_i,
\label{eq:layout_token}
\end{equation}
which encodes the holistic geometric structure of the document, including hierarchical layout patterns and large-scale spatial dependencies.

Finally, $\mathbf{L}$ is projected into the language embedding space and concatenated to the text sequence:
\begin{equation}
\mathbf{X} = \mathtt{Concat}(\mathbf{L}, \mathbf{T}),
\label{eq:layout_concat}
\end{equation}
where $\mathbf{T}$ denotes the text embeddings. 
This allows the language model to jointly attend over textual semantics and globally weighted layout cues, enabling layout-aware and structurally grounded reasoning with negligible computational overhead.


\subsection{Training Recipe}

DocCogito adopts a two-stage training strategy that progressively unifies layout perception, structured reasoning, and reward-driven refinement. 
Stage 1 focuses on learning document-level layout priors through a lightweight Layout Tower, establishing spatially grounded visual representations. 
Stage 2 enhances reasoning through a multi-stage post-training pipeline composed of a VSC-guided cold-start alignment phase, a rejection-sampled supervised finetune phase, and a GRPO reward optimization phase.





\subsubsection{Layout Perception Pretraining}

In stage 1, the process aims to pretrain the layout tower to acquire global layout priors before interacting with the language model. 

Given a document image $\mathbf{I}$ with OCR annotations $\mathcal{B}=\{b_i\}_{i=1}^{M}$, where each $b_i=\langle x_1,y_1,x_2,y_2\rangle$ denotes a text-line bounding box, and M denotes the number of OCR annotations. 
we construct a grid-level supervision map $\mathbf{Y} \in [0,1]^{H \times W}$ by projecting all boxes onto the spatial patch grid $(H,W)$ of the vision encoder. 
The map is normalized to form a valid distribution. 
The layout tower predicts a corresponding spatial distribution $\mathbf{P} \in [0,1]^{H \times W}$ over the same grid.

The pretraining objective encourages $\mathbf{P}$ to match the OCR-derived layout prior using two complementary losses:
\begin{equation}
\mathcal{L}_{\text{total}} 
= \mathcal{L}_{\text{KL}}
+ \lambda_c \mathcal{L}_{\text{center}},
\end{equation}
where $\lambda_c$ is a weighting coefficient that balances the KL loss and the center-alignment loss, and is set to $0.2$.

The first term minimizes the KL divergence between the two grid-level distributions:
\begin{equation}
\mathcal{L}_{\text{KL}}
= \sum_{u,v}
\mathbf{Y}_{u,v}\,
\log \frac{\mathbf{Y}_{u,v}}{\mathbf{P}_{u,v}},
\end{equation}
while the center-alignment loss enforces geometric consistency:
\begin{equation}
\mathcal{L}_{\text{center}}
= \|\mathbf{c}_{\text{pred}} - \mathbf{c}_{\text{gt}}\|_2^2,
\end{equation}
where the predicted and ground-truth centroids are computed as
\begin{equation}
\mathbf{c}_{\text{pred}} = 
\sum_{u,v} \mathbf{P}_{u,v} \cdot (u, v), \qquad
\mathbf{c}_{\text{gt}} = 
\sum_{u,v} \mathbf{Y}_{u,v} \cdot (u, v),
\end{equation}
representing the expectation of spatial coordinates under the predicted distribution $\mathbf{P}$ and the supervision map $\mathbf{Y}$.

This stage equips the layout tower with interpretable and spatially coherent layout-attention patterns, forming a stable structural prior that benefits subsequent reasoning alignment and GRPO optimization.

\subsubsection{Multi-stage Post-training}

After pretraining, the model undergoes a continual three-stage optimization process.

\textbf{Cold Start.}
To stabilize early training steps, we first warm up the policy using the VSC-style CoT dataset described in Sec.~\ref{sec:Dataset Construction}. 
A curated set of 4k structured and layout-grounded samples is used to bootstrap the model, guiding it toward interpretable, step-wise reasoning rather than unconstrained natural-language generation. 
This structure-aware warm-up prevents early-stage policy collapse and ensures that subsequent rejection sampling and GRPO.


\textbf{Rejection Sampling.}
After the above cold start process, we refine the policy on the Multi-Domain QA Corpus introduced in Sec.~\ref{sec:Dataset Construction} through rejection-sampled supervised fine-tuning (SFT). 
For each input, the model generates a single response, which is retained only if it satisfies structural validity and exhibits semantic consistency with the ground truth (\eg, answer-level semantic matching by F1 scores).
Invalid predictions are discarded at this stage and deferred to the GRPO phase, where reinforcement learning further improves the model’s ability to analyze and reason over challenging cases. 
This phase enforces a consistent structured reasoning format and mitigates domain bias by training on diverse yet valid reasoning trajectories.


\textbf{GRPO.}
Following the rejection-sampled fine-tuning stage, we apply GRPO to further refine the policy through reward-driven exploration. 
This reinforcement stage further optimizes the model beyond supervised imitation by sampling multiple reasoning rollouts and updating the policy to maximize structured rewards.
Unlike conventional SFT process, which only mimics reference outputs, GRPO dynamically adjusts the policy based on reward feedback, improving factual correctness, structural and layout grounding, \etal. 
In the next section, we introduce the detailed design of our composite reward function.


\subsection{Reward Formulation}
\label{sec:loss_reward}

To support layout-grounded and structurally consistent reinforcement learning, we design a composite reward with five components:
\begin{equation}
\label{eq:reward_total}
r(x,y) =
r_{\text{ans}}
+ \lambda_q r_{\text{qa}}
+ \lambda_v r_{\text{vsc}}
+ \lambda_s r_{\text{str}}
+ \lambda_r r_{\text{reg}},
\end{equation}
where the auxiliary coefficients are selected via a light hyper-parameter ablation on a held-out validation set and then fixed throughout all experiments:
$\lambda_q=0.20$, $\lambda_v=0.20$, $\lambda_s=0.20$, and $\lambda_r=0.50$.
Rather than relying on delicate tuning, we find the method to be robust to moderate variations of these weights, and we provide comprehensive ablations of both the coefficients and individual reward terms in the supplementary material to substantiate their contributions (details are provided in Suppl.\ Sec.~B).

The reward terms are designed to be \emph{complementary} rather than ad-hoc, each targeting a distinct aspect of the human-like document QA process.
Specifically, $r_{\text{qa}}$ assesses the \texttt{question\_analysis} field, encouraging correct interpretation of question intent and basic planning (\emph{what to do}).
$r_{\text{reg}}$ provides grounding feedback that promotes selecting the appropriate evidence region (\emph{where to look}).
$r_{\text{vsc}}$ evaluates the structural validity of the VSC chain (\emph{how to reason}) by checking schema correctness, operator ordering, region consistency, and basic action diversity, encouraging coherent, layout-referenced step-wise procedures.
$r_{\text{ans}}$ measures task correctness by a composite answer-similarity metric (token-level F1/recall and fuzzy string matching), providing a smooth signal for semantic consistency between prediction and ground truth.
Finally, $r_{\text{str}}$ enforces output-format stability via \texttt{JSON} validity and field-level compliance.

\textbf{Region reward.}
To encourage accurate region grounding in VSC reasoning, we compute a confidence-based reward over region tokens.  
For each VSC step $t$, let $r_t$ denote the region token predicted by the model—the first token following the \texttt{region} field—which specifies the selected layout region:
\begin{equation}
p_t = P_\theta(r_t \mid \mathcal{H}_t, x),
\end{equation}
where $x$ denotes the input document and $\mathcal{H}_t$ denotes the previously generated context.
The region-confidence reward aggregates the log-probabilities of all region tokens:
\begin{equation}
\log r_{\text{reg}} =
\sum_{t=1}^{N} \log p_t,
\end{equation}
where $N$ denotes the number of regions, and then we adopt its length-normalized geometric form as the final reward:
\begin{equation}
\tilde r_{\text{reg}} =
\exp\!\left(
\frac{1}{N}
\sum_{t=1}^{N} \log p_t
\right).
\end{equation}
Since region labels belong to a constrained vocabulary (\eg, header, table, cell, \etal), the first region token provides a stable indicator of region correctness.  
This reward explicitly encourages the model to assign high confidence to correct layout regions, strengthening layout-grounded reasoning in GRPO.


\section{Experiments}

\subsection{Datasets and Metrics}

\textbf{Dataset.} We evaluate DocCogito on a comprehensive suite of text-centric document understanding benchmarks covering structured, tabular, and visually grounded reasoning. 
The benchmarks include DocVQA~\cite{mathew2021docvqa} for multi-paragraph documents with complex layouts, 
ChartQA~\cite{masry2022chartqa} and WTQ~\cite{pasupat2015compositional} for numerical and tabular reasoning, 
InfoVQA~\cite{mathew2022infographicvqa} and TextVQA~\cite{Singh_2019_CVPR} for infographics and natural scenes containing embedded text. 
These datasets assess a model’s ability to read, localize, and reason over textual content under diverse document structures. 
In addition, we adopt OCRBench~\cite{liu2024ocrbench} as a unified benchmark for OCR-based reasoning across heterogeneous document types, offering a holistic evaluation of layout-aware visual–language understanding.

\noindent\textbf{Metrics.} For all benchmarks, we follow the official evaluation metrics, including Exact Match (EM), F1, and task-specific accuracy. 
All results are reported on the official validation or test set as defined by each dataset.




\subsection{Implementation Details}

We adopt Qwen3-VL-Instruct-4B and Qwen3-VL-Instruct-8B~\cite{yang2025qwen3} as our baseline multimodal large language models, 
The overall training and freezing strategy follows the configuration illustrated in Fig.~\ref{figure2}.
We use AdamW throughout training, with stage-specific learning rates: 
$1\times10^{-4}$ for layout tower pretraining, 
$1\times10^{-5}$ for the cold-start stage, 
$5\times10^{-6}$ for rejection-sampled refinement, 
and $1\times10^{-6}$ during GRPO. 
Each stage is trained for 3/3/1/1 epochs, respectively.



\begin{table*}[t]
\centering
\scriptsize
\setlength{\tabcolsep}{2pt}
\resizebox{\linewidth}{!}{%
\begin{tabular}{l|c|ccc|c|c|c}
\toprule
\textbf{\textsc{Model}} & Size(B) & DocVQA & InfoVQA & ChartQA & TextVQA$_{\text{Val}}$ & WTQ & OCRBench \\
\midrule
Donut~\cite{kim2022ocr}  & 0.26 & 67.5 & 11.6 & 41.8 & -- & 18.8 & -- \\
DocFormerv2~\cite{appalaraju2024docformerv2} & 0.75 & 87.8 & 48.8 & -- & 65.6 & 48.3 & -- \\
Mini-Monkey~\cite{huang2024mini} & 2 & 87.4 & 60.1 & 76.5 & 75.7 & -- & 806 \\
HRVDA~\cite{liu2024hrvda} & 7.1 & 72.1 & 43.5 & 67.6 & 73.3 & 31.2 & -- \\
DocKylin~\cite{zhang2025dockylin}  & 7.1 & 77.3 & 46.6 & -- & -- & 32.4 & -- \\
Park et al.~\cite{park2024hierarchical} & 7.2 & 72.7 & 45.9 & 63.3 & 59.2 & 34.5 & -- \\
TextHawk2~\cite{yu2024texthawk2} & 7.4 & 89.6 & 67.8 & 81.4 & 75.1 & 46.2 & 784 \\
TextMonkey~\cite{liu2024textmonkey} & 7.7 & 73.0 & 28.6 & 66.9 & 65.6 & -- & 561 \\
DocOwl2~\cite{hu2025mplug} & 8 & 80.7 & 46.4 & 70.0 & 66.7 & 36.5 & 366 \\
DocLayLLM~\cite{liao2025doclayllm} & 8 & 86.5 & 58.4 & -- & -- & \textbf{58.6} & -- \\
DocOwl-1.5~\cite{hu2024mplug} & 8.1 & 81.6 & 50.4 & 70.5 & 68.8 & 39.8 & 599 \\
DocOwl-1.5-Chat~\cite{hu2024mplug} & 8.1 & 82.2 & 50.7 & 70.2 & 68.6 & 40.6 & -- \\
InternVL2~\cite{chen2024internvl} & 8.1 & 91.6 & 74.8 & \textbf{83.3} & 77.4 & -- & 794 \\
Zhang et al.~\cite{zhang2024token} & 8.1 & 78.3 & 50.2 & 68.9 & 66.6 & 38.6 & -- \\
Marten~\cite{wang2025marten} & 8.1 & 92.0 & 75.2 & 81.7 & 74.4 & 52.4 & \underline{820} \\
Monkey~\cite{li2024monkey} & 9.8 & 66.5 & 36.1 & 65.1 & 67.6 & 25.3 & 514 \\
CogAgent~\cite{hong2024cogagent} & 17.3 & 81.6 & 44.5 & 68.4 & 76.1 & -- & -- \\
\midrule
Qwen3-VL-Instruct~\cite{yang2025qwen3} & 4 & 89.0 & 74.6 & 80.4 & 79.1 & 54.4 & 775 \\
Qwen3-VL-Instruct~\cite{yang2025qwen3} & 8 & 91.6 & 75.7 & 81.3 & 80.7 & 57.1 & 804 \\
DocCogito (Ours) & 4 & \underline{93.1} & \underline{76.0} & \cellcolor{gray!15}82.2 & \underline{81.6} & \cellcolor{gray!15}54.8 & \cellcolor{gray!15}788 \\
DocCogito (Ours) & 8 & \textbf{93.2} & \textbf{78.6} & \cellcolor{gray!15}\underline{82.5} & \textbf{82.4} & \cellcolor{gray!15}\underline{58.3} & \cellcolor{gray!15}\textbf{841} \\
\bottomrule
\end{tabular}
}
\caption{Comparison with representative document-understanding models on multiple benchmarks.
We report model size (in billions of parameters) and performance on six benchmarks, following the officially designated evaluation metrics.
``Val'' denotes the validation set.
Light-gray cells indicate benchmarks whose data type is \emph{not} included in our training mixture (doc/text/inf), i.e., out-of-domain evaluation for our models.}
\label{tab:doc_understanding_subset}
\end{table*}

\subsection{Benchmark Results}

We compare DocCogito with representative multimodal document understanding models and strong Qwen3-VL-Instruct baselines on six benchmarks under official evaluation protocols.
All TextVQA results are reported on the validation set, and OCRBench follows its 1000-point scoring scheme.
For fair comparison, we re-evaluate the baselines under exactly the same metrics and inference protocol as our models (instead of solely relying on officially reported numbers); details are provided in Suppl.\ Sec.~C.

As shown in Table~\ref{tab:doc_understanding_subset}, both our 4B and 8B variants achieve strong and consistent performance across all tasks. The 8B model achieves new state-of-the-art results on DocVQA, InfoVQA, TextVQA$_\text{val}$, and OCRBench, surpassing the previous SOTA method by 1.2, 3.4, 5.0, and 2.1 points, respectively. These gains suggest that the proposed layout-aware architecture and multi-stage training recipe jointly enhance text-centric perception and reasoning over visually structured documents.

\begin{figure}[t]
\centering
\begin{minipage}[t]{0.49\linewidth}
  \centering
  \includegraphics[width=\linewidth]{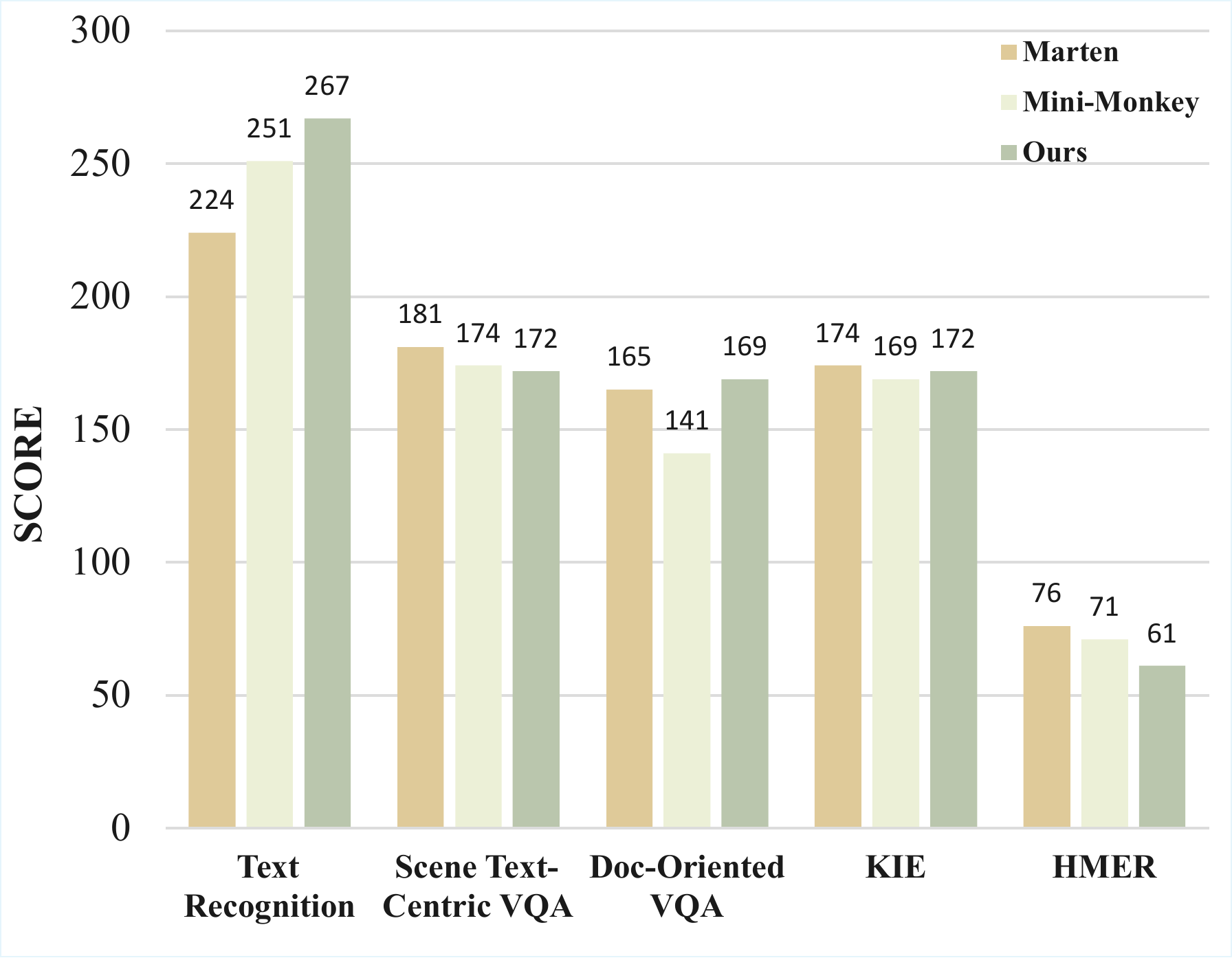}
  \caption{OCRBench subtask breakdown for the top three overall models (Marten, Mini-Monkey, and DocCogito-8B) under the official evaluation protocol.}
  \label{fig:chart}
\end{minipage}\hfill
\begin{minipage}[t]{0.49\linewidth}
  \centering
  \includegraphics[width=\linewidth]{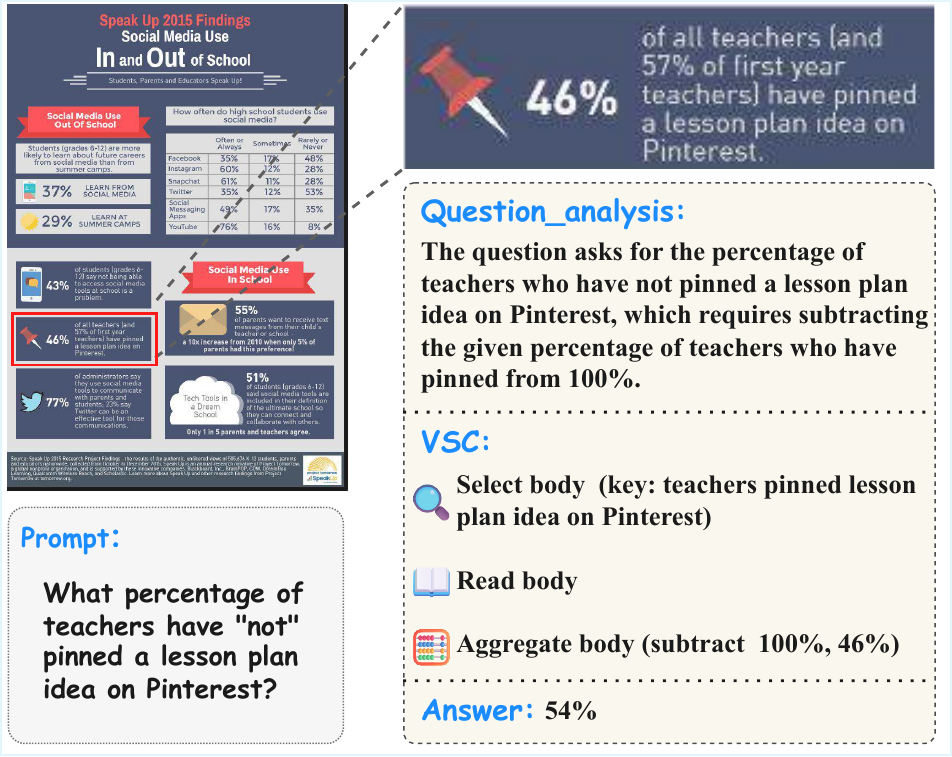}
  \caption{A VSC-style CoT example showing question analysis, region grounding, and operator-level reasoning steps that lead to the final answer.}
  \label{fig:vis}
\end{minipage}
\end{figure}

A breakdown of OCRBench in Figure~\ref{fig:chart} further shows that the improvement is broadly distributed across all five subtasks rather than concentrated in a single component. The 8B model leads by a large margin in text recognition and remains competitive on scene text-centric and doc-oriented VQA, indicating robustness from fine-grained token perception to higher-level semantic reasoning. Steady gains on KIE and HMER additionally confirm that DocCogito strengthens structured grounding and symbol-level understanding, yielding a balanced and holistic OCR capability.

Beyond these results, the 8B model also achieves comparable performance on ChartQA and WTQ, closely approaching the leading methods. The 4B variant, despite having substantially fewer parameters, remains highly competitive and ranks on DocVQA, InfoVQA, and TextVQA—surpassing several 8B–17B models. This demonstrates the parameter efficiency of our framework and its ability to deliver strong grounding and structured reasoning even at smaller scales.

Compared with the Qwen3-VL-Instruct baselines, DocCogito delivers consistent improvements across all six benchmarks at both model scales.
Notably, the gains persist on benchmarks whose data types are not included in our training mixture (gray cells), indicating that the improvements are not merely due to in-domain memorization.
These consistent in-domain and out-of-domain gains strongly validate the design of our layout-grounded reasoning paradigm.

\subsection{Ablation Study}


\textbf{Components of DocCogito.}
We conduct ablations to evaluate the contributions of the VSC-style CoT, the layout tower and the GRPO training stage, using the 4B model under the standard setting. Results are presented in Table~\ref{tab:ablationr}.

Removing the VSC-style CoT leads to consistent performance drops across all benchmarks, 
most notably on TextVQA ($-2.4$). 
where spatial grounding and step-wise decomposition are essential. 
The VSC formulation provides deterministic and region-referenced supervision, 
enabling the model to execute more stable and layout-aware reasoning.

Removing the layout tower leads to consistent performance degradation across all benchmarks, with drops of 1.6, 1.0, 2.6, and 1.9 points on InfoVQA, TextVQA, ChartQA, and WTQ, respectively. This confirms that explicit global layout encoding provides complementary structural cues that enhance grounding and question-specific reasoning. Eliminating GRPO results in larger declines, particularly on reasoning-centric tasks such as TextVQA and WTQ, indicating that reinforcement learning plays a critical role in refining multi-step reasoning and response calibration beyond supervised fine-tuning.

\begin{table}[t]
\centering
\setlength{\tabcolsep}{3pt}
\renewcommand{\arraystretch}{0.95}

\begin{minipage}[t]{0.49\linewidth}
\centering
\captionof{table}{Ablation study of key components in DocCogito.}
\label{tab:ablationr}
\scriptsize
\resizebox{\linewidth}{!}{%
\begin{tabular}{lcccc}
\toprule
Method & InfoVQA & TextVQA & ChartQA & WTQ \\
\midrule
DocCogito-4B & \textbf{76.0} & \textbf{81.6} & \textbf{82.2} & \textbf{54.8}\\
\midrule
w/o VSC-style CoT & 74.9 & 79.2 & 81.0  & 53.5\\
w/o layout tower & 74.4 & 80.6 & 79.6 & 52.9\\
w/o GRPO & 71.2 & 77.5 & 79.8 & 52.6\\
w/o layout tower \& GRPO & 71.8 & 77.1 & 80.8 & 53.4\\
\bottomrule
\end{tabular}%
}
\end{minipage}\hfill
\begin{minipage}[t]{0.49\linewidth}
\centering
\captionof{table}{Ablation across different backbones and model sizes.}
\label{tab:basemodel}
\scriptsize
\resizebox{\linewidth}{!}{%
\begin{tabular}{lcccc}
\toprule
Base Model & InfoVQA & TextVQA & ChartQA & WTQ\\
\midrule
Qwen3VL-8B~\cite{yang2025qwen3} & \textbf{78.6} & \textbf{82.4} & \textbf{82.5} & \textbf{58.3}\\
Qwen3VL-4B~\cite{yang2025qwen3} & 76.0 & 81.6 & 82.2 & 54.8\\
Qwen2.5VL-3B~\cite{Qwen2.5-VL} & 69.1 & 77.9 & 82.2 & 51.7\\
\bottomrule
\end{tabular}%
}
\end{minipage}

\end{table}

When both layout tower and GRPO training stage are removed, performance deteriorates further, suggesting that layout-aware representation and reasoning-oriented optimization contribute independently and are jointly necessary for strong document-understanding capability. These findings demonstrate that DocCogito’s improvements stem from both architectural and training-stage innovations rather than parameter scaling alone.

\noindent\textbf{Baseline models.}
We further evaluate whether DocCogito’s improvements arise from the proposed framework rather than the choice of backbone or model size. Table~\ref{tab:basemodel} reports results using Qwen3VL-8B, Qwen3VL-4B, and Qwen2.5VL-3B as the underlying models.

The scaling law is very evident. Across all four benchmarks, larger base models yield consistently higher performance, demonstrating that DocCogito scales effectively with model capacity. More importantly, the relative improvements remain stable across backbones: the 4B variant achieves performance close to the 8B model on ChartQA (82.2 vs. 82.5) and TextVQA (81.6 vs. 82.4), and the 3B model also attains competitive results, particularly on ChartQA. This consistency indicates that the framework provides benefits independent of model size and generalizes well to both small and large vision–language architectures.


\begin{figure}[t]
    \centering
    \includegraphics[width=1\linewidth]{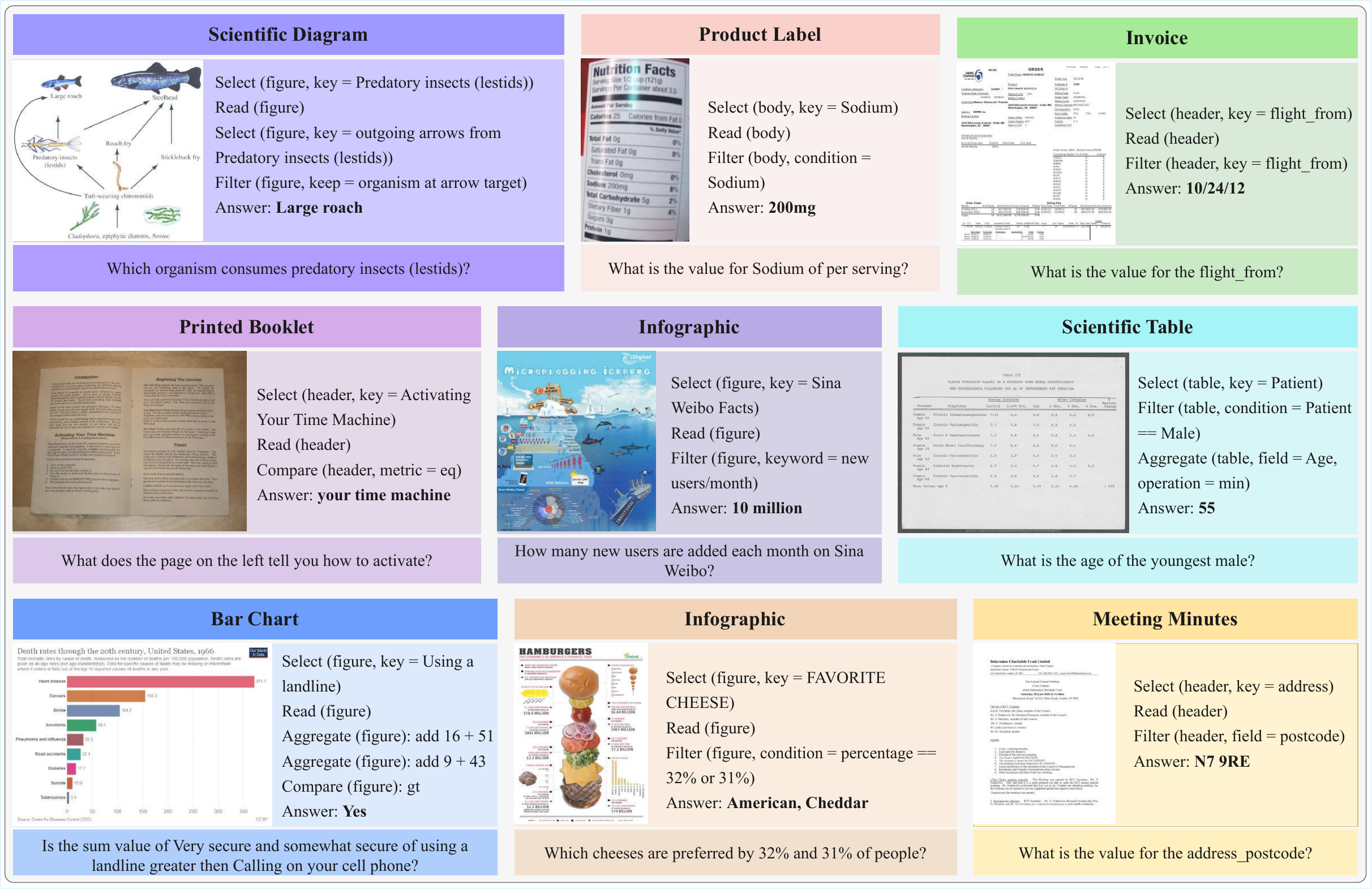}
    \caption{Qualitative examples of our VSC-style CoT across diverse document types.}
    \label{fig:vis_all}
\end{figure}
\section{Visualization and Analysis}

Figure~\ref{fig:vis} presents a representative VSC-style CoT reasoning trace produced by DocCogito.
Given a document image and a natural-language query, the model first generates \texttt{question\_analysis} to clarify the intent and identify the required operation pattern.
It then executes a VSC sequence with explicit, region-grounded steps: \textbf{Select} localizes the evidence region, \textbf{Read} extracts the relevant textual content, and \textbf{Aggregate} performs a deterministic computation to obtain the final answer.

Beyond this single example, Fig.~\ref{fig:vis_all} provides a diverse collection of VSC-style CoT cases spanning multiple document types and question forms.
These examples serve as an initial empirical validation that our compact operator set can express the reasoning patterns required by the current evaluation suite, while maintaining explicit region references throughout multi-step inference.

Overall, the visualizations highlight the advantages of the VSC paradigm.
Compared with free-form CoT, VSC enforces a structured, region-referenced reasoning process driven by deterministic operators, improving interpretability and exposing how DocCogito transitions from layout perception to step-wise semantic reasoning.
The consistent grounding patterns observed across cases further suggest that our progressive post-training pipeline encourages stable, layout-faithful reasoning behavior.


\section{Conclusion}
We presented \textbf{DocCogito}, an OCR-free document understanding framework that explicitly couples global layout cognition with step-level, region-grounded reasoning. DocCogito distills page structure into a lightweight \textbf{layout tower} that produces global layout prior tokens, and aligns intermediate inference with evidence regions through a deterministic \textbf{Visual--Semantic Chain (VSC)}. Building on this structured interface, we further developed a progressive post-training pipeline---from layout perception pretraining and VSC-guided cold start to rejection sampling and GRPO---augmented with a fine-grained region-confidence signal to strengthen layout--reasoning coupling.

Extensive experiments across six benchmarks demonstrate consistent gains over strong baselines and state-of-the-art performance on multiple tasks, while ablations and qualitative analyses verify that both the layout prior and the structured, region-referenced execution are necessary for reliable document reasoning. We hope DocCogito encourages future work toward more auditable and grounded document MLLMs, including extending the operator set to richer actions, scaling to multi-page settings, and improving robustness under severe layout shifts and long-horizon evidence aggregation.

\bibliographystyle{splncs04}
\bibliography{main}

\end{document}